# Swa Bhasha: Message-Based Singlish to Sinhala Transliteration

Maneesha U. Athukorala
School of Computing
Informatics Institute of Technology
Ramakrishna Road, Colombo 006, Sri Lanka
maneeshaathu4@gmail.com

Deshan K. Sumanathilaka
School of Computing
Informatics Institute of Technology
Ramakrishna Road, Colombo 006, Sri Lanka
deshan.s@iit.ac.lk

*Abstract*—Machine Transliteration provides the ability to transliterate a basic language into different languages in a computational way. Transliteration is an important technical process that has caught the attention most recently. The Sinhala transliteration has many constraints because of the insufficiency of resources in the Sinhala language. Due to these limitations, Sinhala Transliteration is highly complex and time-consuming. Therefore, the majority of the Sri Lankans uses non-formal texting language named 'Singlish' to make that process simple. This study has focused on the transliteration of the Singlish language at the word level by reducing the complication in the transliteration. A new approach of coding system has invented with the rule-based approach that can map the matching Sinhala words even without the vowels. Various typing patterns were collected by different communities for this. The collected data have analyzed with every Sinhala character and unique Singlish patterns related to them were generated. The system has introduced a newly initiated numeric coding system to use with the Singlish letters by matching with the recognized typing patterns. For the mapping process, fuzzy logic-based implementation has used. A codified dictionary has also implemented including unique numeric values. In this system, Each Romanized English letter was assigned with a unique numeric code that can construct a unique pattern for each word. The system can identify the most relevant Sinhala word that matches with the pattern of the Singlish word or it gives the most related word suggestions. For example, the word "kiyanna,kianna, kynna, kynn, kiynna" have mapped with the accurate Sinhala word "කියන්න". The system has presented 84% of accuracy in word-level and 92% of accuracy in suggestion-level prediction. These results revealed that the "Swa Bhasha" transliteration system has the ability to enhance the Sinhala users' experience while conducting the texting in Singlish to Sinhala.

*Keywords*—*rule-based machine transliteration, singlish words, sinhala transliteration, swa bhasha*

## I. Introduction

Language transliteration from one language into a different language is the most common and popular method which is used all over the world. Starting from using human translators for the language translation process, today with the revolution in technology, people can use simple software to achieve their language transliteration needs. However, Human transliteration is the best transliteration method that every country uses even today [1].

Sri Lanka is a South Asian country that has around 21,436,802 people population [2]. Among this population, approximately 16 million people [3] use 'Sinhala' as one of the national languages in Sri Lanka as their native language. The Sinhala language is related to the Indo-Aryan language family which is a branch of the Indo-European languages [4]. Transliteration in the Sinhala language context is challenging, because of the alphabet in the Sinhala language.

In the Sinhala alphabet, there are 54 basic characters with 18 vowel letters, 36 constant letters. 18 vowel letters have included 8 stops, fricatives 2, affricatives 2, nasals 2, liquids 2 and glides 2 [4]. Many systems have been implemented to transliterate into the Sinhala language. For the transliteration process, NLP (Natural Language Processing) techniques have been used. Most of the time MT (Machine Translation) a subset of NLP has been used for the language transliterations with various algorithms using various approaches. In MT, systems can be classified into seven system categories which can define as Rule-based (RB), Statistical, Example-based (EB), Human-assisted, Knowledge-based, Hybrid, Dictionary, and Agent-based [5].

When addressing the transliteration process in texting, most people like to use an easy language technique while typing. Sri Lankan native Sinhala users, especially the new generation use 'Singlish' an informal texting language that mixes both English and Sinhala while texting. But, using 'Singlish' they can get the Sinhala words in English letters or they have to input the Singlish word with vowels. If they need to get the output in native Sinhala, they have to use a Sinhala letter keyboard.

To satisfy this today, people can use the 'Helakuru' keyboard [6]. This is the most suitable and commonly used keyboard in Sri Lanka, which uses native Sinhala while texting. It is really helpful and easy to use. But on this keyboard, users have to use the Singlish word with vowels. Then the user can get the correct word in Sinhala. But if users typed any Singlish word even without vowels, that system will not give the correct Sinhala word as the output. But they will suggest the Sinhala word if you use that Sinhala word frequently. Therefore, finding a unique way to use Singlish and get the output in native Sinhala is essential and challenging. Because people's typing patterns are different from each other.

In this paper, we present a system that can transliterate Singlish words even with vowels or without vowels into native Sinhala at the word level. A novel rule-based mapping system has been introduced in the system to support Singlish words without vowels transliteration into native Sinhala along with the NLP techniques and fuzzy logic theories.

## II. Literature review

### A. Sinhala Language in Machine Transliteration

The English-Sinhala Decoder [7] invented in 2018 was the latest modern bilingual translator. This has been used to translate business-related documents in Sri Lankan organizations to ensure more confidence in dealing with the international and minimize the errors made by the human translators. In this translator, the creators have especially addressed for sections to develop the Sinhala-English translator, implement Singlish transliteration and Sinhala handwritten recognizer. For this Sinhala- English bilingual translator, creators have been used the Google API, a statistical-based approach to translate the input sentences and verify or edit their semantic meanings to match with the Sinhala language using rule-based, knowledge-based and example-based approaches. In this system, the rule-based approach has a separate database with Sinhala and English definition words dictionary with 70,000 wordlist and rules to create meaningful sentences. In this rule-based process, the source sentence (input sentence) will be split into separate words and searched for morphological information using the database. The example-based approach will be compared to the source sentence with the database that includes 100,000 example sentences and postfixes the sentence. The knowledge-based approach will provide synonyms for the targeted sentence by using the database which includes 50,000 words. In this method also the target sentence will be split into words to find and match the suitable synonyms. The 'Singlish translator' has implemented to do the transliteration under three segments starting from manipulating the entered English words by creating a list of related English characters. Then the generated string characters will pass to the phonemic rule base and a transliteration rule base related to the Sinhala alphabet. After these processes are completed, the system will generate a list of suggested words for that particular input transliterated word. The Sinhala handwritten recognizer will recognize Sinhala letters according to three vertical positions named as upper, middle and lower zones and identify a total of 42 characters in the Sinhala alphabet. Then, each letter will separate into five categories. When the recognizing process has completed for each character, the system has been generated a probability percentage for each character's accuracy and provided a suggestion for the next most suitable matching character.

The Office Documents MT-based system for Sinhala and Tamil [8] discovered in 2018 was the first official document translator for both Sinhala-Tamil languages. This system was specially addressed in the government sector. In this system, the researchers have found that the capability of translating the source language documents which is the Sinhala language into the target language (Tamil) was high compared to other translators. And also, the lack of a post-editing facility in the existing systems was covered by this 'Si-Ta' system. The current version of 'Si-Ta' has the ability to translate documents with up to two pages and most importantly this system is a bilingual translator. This system starts when the source document is input into the system. Then the system highlighted words that cannot translate at the system level. A manual translation needs to be conducted for those highlighted words in the document. The system has been implemented using the semantic approach.

The Trigram and Rule-based module [9] invented by W.M.P. Liwera and L. Ranathunga, is a modern approach that has implemented to transliterate Singlish to Sinhala. This module has been highly focused on social media texting. The rule-based approach has used to transliterate Singlish words automatically but has modified by combining together with and trigram method to avoid the confusion that occurred during taking the correct results from the system. The Singlish transliteration schema that has implemented by the University of Colombo has used in this system as the rule-based approach to perform Unicode transliteration support to the Singlish. This schema has operated upon a set of rules that have separately defined as 46 rules for the comparisons of consonants, 26 rules for comparisons of 14 vowels and 8 rules for special characters and consonants related to the Sinhala language. The inventors have identified that many different typing patterns have used in the social media platforms and modified some existing rules in the UCSC transliteration schema to get more correct and better results. They have used the Trigram tagger to get a better accuracy to the system by using the NLTK library. All three unigram, bigram and trigram have used in the transliteration process. This Trigram has also used in the training process of the translator. This Trigram and Rule-based approach combined system module has given an overall better accuracy as 62% of a word-level accuracy and 77% of letter-level accuracy.

Pali-Sinhala Machine Translation [5] is a translator that can used as a self-learning tool also when translating Pali into the Sinhala Language. This system can translate simple Pali sentences to the Sinhala language by using a dictionary-based approach. This System has consisted of three models named Pali morphology analyzer, dictionary-based translator and the Sinhala morphology generator. The Pali morphological generator was the main section in this system. Because it will read the input sentence at word (Pali) level and provide the root word and the relevant grammatical information related to those words using the affix spiriting approach. Root word table has been used for the regular Pali words and the irregular words have been directed to the Pali dictionary. In terms of Pali nouns, this language is normally supported for gender-based conjugation. Therefore, for this system, the inventors had been identified an affix spiriting table for nouns and verbs in the Pali language to identify the root word, using the method of dividing the conjugation into the last sound of the noun named 'Anthaya'. The Pali lexical database that contains irregular and regular Pali nouns and verbs have been connected to fulfil this process. Pali-Sinhala dictionary has been used for identifying the base Sinhala word for the input Pali word. In this model, a Pali-Sinhala bilingual dictionary has been included with Pali words and their related grammatical categories with Sinhala words. The last and final model in this system is the Sinhala morphology generator. The use of this model is to generate the relevant Sinhala word to the base word identified from the Pali-Sinhala dictionary model using the affix spiriting rules in the system. In this system, a limited number of rules have been used since the inventors had used the Pali sentences in the Dhamma school's textbooks. Afterwards, the translated Sinhala words have been displayed as the output. The word order of the resulting sentence is sorted since the Pali and Sinhala word orders are normally related for both languages.

Sinhala-English Machine Translation [3] is a translator that can translate grammatically correct Sinhala sentences into relevant English sentences. The features that specific for this system was, this has been the first system in Sri Lanka to have features such as an inbuilt dictionary, keyboard, Unicode fonts based integrated word processor, debugger, grammar tool, Sinhala grammar checker and a word tool. To build this system, creators have used a rule-based approach named 'the transfer-based machine translation approach'. And also, the object-oriented programming methodology to design the system. For the implementation, they have been used the visual studio .Net framework with visual basic .Net, and database management system- Microsoft access 2003. The processes of the system will be started with the Sinhala grammatically correct sentences or a paragraph that is input by the user. The paragraph or the sentence should be separated with relevant full stops and other punctuation characters because the system uses them to identify a full sentence when using full stop or a question mark and space, comma, colon, semicolon identify as the end of a selected word while analyzing. After identifying a full sentence, each word in that sentence will be added to a separate object. This object will be added into an ArrayList and it will continue until the analyzing process ends. Then the translation engine identifies whether the input sentence or paragraph is English or Sinhala. At the end of that process, the sentences will be added to a data table. In this data table, all the information related to each word will be included. Each sentence will have a separate data table, and in this table, each word in the sentence will have a separate row. In this process, the translator engine can combine suitable words according to the grammar rules. Then the string pattern creation will be started for the Sinhala sentence that can use for the English string pattern later. Then the word formatting will be started, starting from the subject and then the verb will be formatted according to its tense, and voice. Then these all the words will be added into a string. At the end of the translation process, all the words that are formatted and do not necessarily need to be formatted will be connected into the same string according to the sentence pattern (English) and end the translation by giving the correct output in English.

*B. Different Languages in Machine Transliteration*

The Hindi-English Translator [10] is a translator which can help to translate the Hindi language into English at the phrase level. This system is a special approach because the implementers have used a hybrid approach to implement this by combining statistical, rule-based and example-based machine translation approaches. For the implementation of this system, the implementers have been used four main primary steps named segmentation, translation. POS tagging, and rearrangement. This system starts when the user input some sentences, or a sentence typed in the Hindi language. Then the segmentation part starts, and the system will separate those sentences or the sentence into words. This system has a database also. While doing this sentence separation if the system found that a particular word is example-based, then it will directly be converted into English. The purpose of this segmentation is to identify possible words that are related to the Hindi-English database. While the segmentation is ongoing, the POS tagging also will starts. After the tagging is done, the translation process will happen at the word level or the sentence level. The statistic-based translation will help to prevent the multi meanings problem of the segmented words which means ambiguity. This translation will process according to the pos tagging and that will identify similar words that have already existed in the Hindi-English database. In the final stage, the rearrangement stage will do the rearrangement of the translated words which have been segmented earlier in the sentence according to the grammatical rules with the support of the rule-based machine translation approach and produce a correct sentence. Finally, the proper output will give in the English language.

The English-Hindi Machine Translation [11] system is a translating system that can use to translate English into Hindi effectively. This translator has been invented as a solution for those who have a lack of knowledge in the English language domain in India. This is also a hybrid approach that has invented combing rule-based, and statistical-based machine translation approaches. In this system, the processers will start with the input sentence or sentences in the English language. After that, the splitter immediately divides the English sentence or the sentences into a set of words. After that, that set passes to the parser. The Stanford parser will process through the word set and analyze the grammatical structure of the English sentence. POS or 'Part of Speech' tagger will tag the words in the word set according to a 'trained' pre-created file including the information about different tags. These tags will generate information about the word set related to the inputted sentence. In this system, there are special declension rules which have been created according to the grammatical rules in the Hindi language. This will be tagged the POS tags in the word set including those grammatical rules. There is a separate sentence file has been included with the relevant rules to translate English sentences into Hindi sentences. Then the reordering part will reorder the sentences according to the pre-defined rules. This step is essential because the sentence pattern in English is different from the Hindi sentence pattern. At the end of the reordering, the translator will translate the sentences into Hindi with the help of the lexicon database and give the correct translated output.

Gurmukhi-Roman Transliteration system [12] is a character-mapping and hand-crafted rules-based system. The system has implemented to use in many areas such as, in the normalization process of the social media texts, summarization and analysis of the multilingual texts and also in translation processes. The character mapping has done using the Gurmukhi scripts. This system has mainly transliterated the Punjabi text which has written using Gurmukhi scripts into English language text. In this system, the inventors have conducted a hybrid approach by combining the rule-based and character-based approaches. The GRT system has shown a 99.27% accuracy, which means a better accuracy after testing approximately 65,130 various part-of-words related to the Panjabi language as the result.

By considering all the above researches we have decided to proceed the research with the rule-based approach along with the NLP techniques.

III. METHODOLOGY

We have selected the structured system analysis development methodology (SSADM) as the system design methodology because, the proposed system will be not required to contain any classes since the project is based on the data science approaches, and the system will be done in a sequence-to-sequence manner. Therefore, each previous stage in the system has to be completed to go to the next step in the system.

The implementation work has been carried out using the Jupiter Notebook provided by Anaconda Navigator environment with Python as the Programing language. We have selected 'Tkinter' as the front-end GUI framework. Natural Language Toolkit (NLTK) library has been used for the NLP, MT-based developments in the system and fuzzy-wuzzy library has been used for the data matching process of the system with the user inputs.Fig.2 shows the system diagram of the Swa Bhasha transliterator.

The data set holds a significant part in this data science project. Therefore, a dataset containing native Sinhala words which are currently used by the Sri Lankans have been used in the system. We have done some changes such as giving unique numbers for each native Sinhala word to the dataset in order to use it in the implementation process. The processes of the 'Swa Bhasha' system's core functions have mentioned below.

### A. Process the Input word, tokenization and the process in the letter dictionary

The system starts the process when the user inputs the Singlish word into the system. Then the system will send the word for the tokenization process to divide the word into simple letters. Afterwards, the tokenized words will add to a list(list one). This list sends to the letter dictionary. In this letter dictionary, the entire English alphabet has included with a unique integer which has assigned for each letter. The list with the input word's letters (list one) then compares with this dictionary and identify whether the letters in the list are in the dictionary as keys. Then the identified dictionary keys and assigned values for that particular letter list will send to separate list (list two). This list(list two) holds both keys which means the letter and the value as tuples. Therefore, the values separate from the current list and send to another new list(list three) according to the exact letter order that the user has input at the beginning of the process. Afterwards, the list of the letter values (list three) will enter into the mapping process.

### B. Rule mapping and the matching with the dataset

In the initial stage of the mapping process, a condition has been added to check whether the list(list three) in the previous stage holds values greater than 10. The purpose of this step is to identify whether the vowels are located in this list or not. If the condition is true, then the system identifies this list as a list of constant letters, which means this word is a 'word without vowels'. Then the system calculates the total letter count in that list (list three). After that, the system will be sending that list into separate conditions according to the total count of the list elements. These conditions have been created from 1 to 9. In this, each of the conditions has separate lists with indexers that vowels have located. These indexes have pre-defined in the system with the relevant positions that vowels can be located in Singlish words. These indexes have been created by studying the different vowel combinations in the Sinhala words. Fig.1 illustrates the vowel combination of the word 'kohomada'. For each position, there will be a separate list. Therefore, the values in the current list(list three) will be added to a new list(list four) and the pre-defined values which have been created according to different vowel values will be inserted into the same list. Each list represents a separate word. Afterwards, the values in each list (list four) will be merged and created a value. Fig.3 illustrates the transliteration process of the Singlish word 'khmd'(kohomada) in a without vowels scenario.

If the list three holds any values greater than 10, then the system identifies it as 'words with vowels' or 'words with the mixed condition', which means a word that reduces the correct vowel count in the real world. After that, the letters list(list three) will send to the value matching process. Then this merged value connects with the native Sinhala dataset using the 'fuzzy-wuzzy' library. This library will match the merged value with the values in the native Sinhala words dataset. As the next step, the identified matched value or values will be generating the result and finally, displayed in native Sinhala.

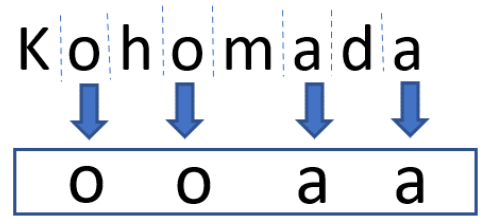

Fig. 1. The vowel combination of the word 'kohomada'.

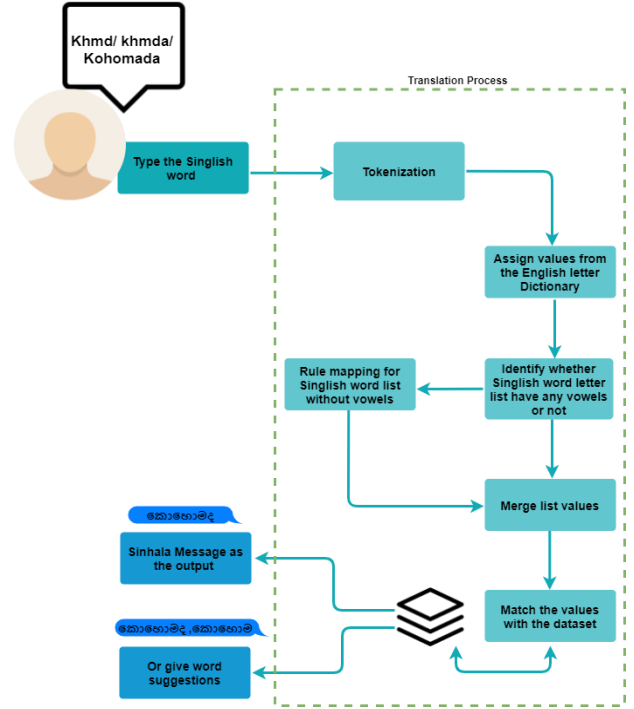

Fig. 2. The system workflow of the Swa Bhasha transliterator.

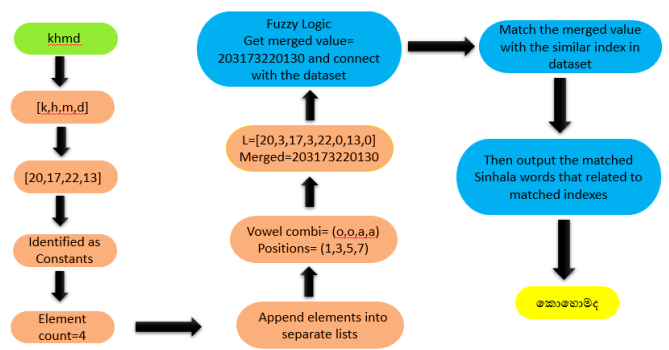

Fig. 3. Example for the process of Singlish words without vowels transliteration.

## IV. EXPERIMENTAL RESULTS AND EVALUATION

The testing of this system has been conducted to address all three transliteration scenarios such as,

- Singlish transliteration without vowels
- Singlish transliteration with vowels
- Singlish transliteration of the Singlish words that have reduced the vowel count in the real Sinhala word

Swa Bhasha system's functional requirement testing was conducted using the Black box testing approach. This black box testing is used to validate the user's inputs with the system's outputs.

The aim of conducting this testing process was to validate that the above-mentioned scenarios are functioning well with better accuracy because this is a system that the outputs will be entirely dependent on the user's inputs. The current system function in word-level Singlish transliteration and it gives the relevant Sinhala word or word suggestions as the output.

*A. Accuracy Testing for Singlish words without vowels*

The accuracy test has conducted by creating test cases to cover up the mapping conditions using the words in the native Sinhala dataset. Each test case represents a native Sinhala word in the dataset related to a mapping condition. These mapping conditions have been created after considering the vowel patterns of the Sinhala words and their positions in Sinhala words. TABLE I shows some of the vowel patterns of the Sinhala words that have been used for the testing process and they have been used to find the accuracy of the words without vowels. Approximately 30 vowel patterns related test cases have used to test the accuracy of the system along with the "Google input tool" and "Helakur" application. Among the test cases, 21 test cases have successfully given the accurate output by the "Swa Bhasha" system as a correct single output or with the suggestions. And most importantly, this feature in the "Swa Bhasha" system has given successful results compared to the Google input tool and the helakuru application.

*B. Accuracy Testing for Singlish words with vowels and minimizing the vowel count of the real Sinhala word*

To find the accuracy of the words with vowels and the words with the minimized vowel count, we have used the words that are frequently used by Sri Lankan people which have been collected from the targeted audience and the words from the native Sinhala dataset. TABLE II shows some of the Singlish words with their Sinhala native words that have been used for the testing process. Approximately 50 words have tested under this scenario and more than 40 words have given successful results as single or suggestion level outputs.

According to the total results of the test cases related to the mapping rules and the words taken from the dataset the accuracy level showed a better level as an 84% success rate for word-level transliteration and a 92% success rate for suggestion level predictions with the tested data. Fig.4, Fig.5 and Fig.6 illustrates the interfaces of the system according to the three scenarios with the example results. According to these results, we can conclude that the Swa Bhasha Singlish to Sinhala transliteration performs well with each core feature.

TABLE I. VOWEL COMBINATIONS

| Vowel Patterns | Related Word/s |
|---|---|
| a,a | Baha (බැහැ), Thaththa (තාත්තා) |
| a,u | Pasu (පසු), Kalu (කළු) |
| a,a,a | Dawasa (දවස) |
| o,a,a | Mokada (මොකද) |
| a,i,a,a | Hariyata (හරියට) |
| i,a,a,a | Nidaganna (නිදාගන්න), Nidahasa (නිදහස) |
| i,e,a,a | Thibenawa (තිබෙනවා) |
| o,o,a,a | Kohomada (කොහොමද) |

TABLE II. TEST DATA FOR SINGLISH WORDS WITH VOWELS AND REDUCE THE VOWEL COUNT IN THE REAL SINHALA WORD SCENARIOS

| Singlish Word | Sinhala Word |
|---|---|
| Kiyanna | කියන්න |
| Innawa | ඉන්නවා |
| Kohomada | කොහොමද |
| Karanne | කරන්නේ |
| Amma | අම්මා |
| Ethakota | එතකොට |
| Neda | නේද |
| Wahanse | වහන්සේ |
| Dunna | දුන්නා |
| Epa | එපා |

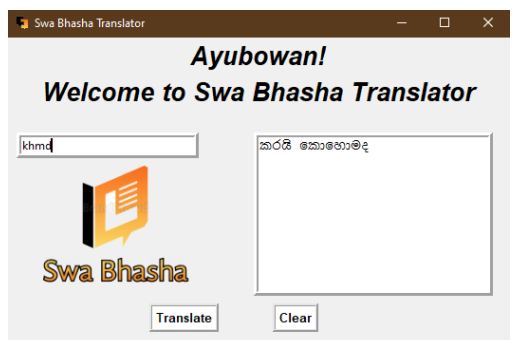

Fig. 4. System UI example 'khmd'(kohomada) for Singlish words without vowels transliteration scenario.

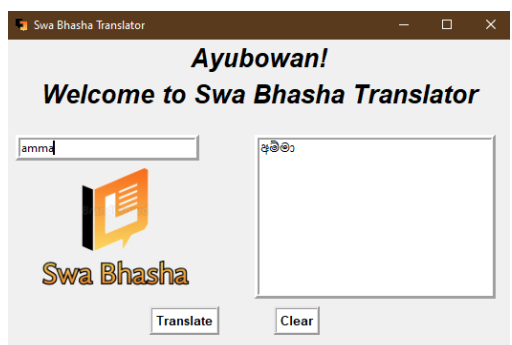

Fig. 5. System UI example 'amma' for Singlish words with vowels transliteration scenario.

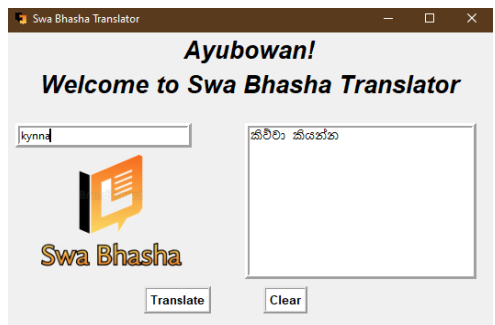

Fig. 6. System UI example 'kynna'(kiyanna) for Singlish words that reduced the vowel count of the real Sinhala word transliteration scenario.

## V. LIMITATIONS AND FUTURE WORKS

"Swa Bhasha" Singlish to Sinhala transliteration system has developed to transliterate Singlish words, which have been used in everyday life into native Sinhala words. Therefore, any grammatical rules have not been included in the current system. This system is currently addressed only at the word level. A limited data set has used in the system.

In this project, we have implemented a web-based system as the prototype. Therefore, this prototype can be enhanced as a mobile application. In addition, the mobile application also can be enhanced using a suitable keyboard that supports Sinhala typing. The data automation process is also a feature that needs to be added as a future enhancement. And also, the current system has developed without concentrating on grammar and sentence transliteration. Therefore, as a future enhancement, the sentence and paragraph level transliteration with Sinhala grammar rules can be included in the system.

## VI. CONCLUTION

The message-based Singlish to Sinhala transliteration is a challenging research area in terms of the Sinhala language because Sinhala is known as the low resourced language. Although several types of research have been conducted and many transliteration applications are existing related to the Sinhala language context but it seems to be that the main objective of those researches and applications was to implement a system or an application for transliterating the Sinhala language into English, different language or to use as an official document transliterator. Although these systems can be helpful in many perspectives under different scenarios, modern texting using Singlish words is can be difficult because the typing patterns of the people are really difficult to understand. Therefore, it is important to pay the attention to people's typing patterns and try to provide a system that can translate Singlish and give the native Sinhala output according to their typing patterns. "Swa Bhasha" system can be a better solution to overcome the current Singlish typing matters.

## ACKNOWLEDGMENT

The author would like to express gratitude to the author's parents and to the author's supervisor for providing the greatest support and guidance throughout the project. The author would like to thank Dr Ruvan Weerasinghe for providing the data set that was needed to develop the system. And also, the author would like to thank all the reviewers who have reviewed this research paper and added their valuable suggestions and comments. Finally, the author thanks everyone that has contributed to making this research successful.

## REFERENCES

[1] B. Hettige, A. Karunananda and G. Rzevski, "Phrase-level English to Sinhala machine translation with multi-agent approach," in 2017 IEEE International Conference on Industrial and Information Systems (ICIIS), Peradeniya, Sri Lanka, 2017.

[2] Worldometers, "Sri Lanka Population," 2020. [Online]. Available: https://www.worldometers.info/world-population/sri-lanka-population/.

[3] A. A. I. U. V. K. D. K. P. S. T. Dilshan De Silva, "Sinhala to English Language Translator," in 2008 4th International Conference on Information and Automation for Sustainability, Colombo, Sri Lanka, 2008.

[4] S. Gallege, "Analysis of Sinhala Using Natural Language Processing Techniques," in Analysis of Sinhala Using Natural Language Processing Techniques , 2010.

[5] B. H. R. M. M. Shalini, "Dictionary Based Machine Translation," in Dictionary Based Machine Translation, Moratuwa, Sri Lanka, 2017.

[6] Helakuru, "හෙළකුරු," Helakuru, [Online]. Available: https://www.helakuru.lk/.

[7] A. I. S. S. C. P. S. T. A.J. Vidanaralage, "Sinhala Language Decoder," in 2018 National Information Technology Conference (NITC), Colombo, Sri Lanka, 2018.

[8] F. F. U. T. S. J. G. D. Surangika Ranathunga, "Si-Ta: Machine Translation of Sinhala and Tamil Official Documents," in 2018 National Information Technology Conference (NITC), Colombo, Sri Lanka, 2018.

[9] L. R. W.M.P. Liwera, "Combination of Trigram and Rule-based Model for Singlish to Sinhala Transliteration by Focusing Social Media Text," in IEEE, Colombo, Sri Lanka, 2020.

[10] S. M. U. S. T. Omkar Dhariya, "A Hybrid Approach For Hindi-English Machine Translation," in 2017 International Conference on Information Networking (ICOIN), Da Nang, Vietnam, 2017.

[11] A. K. K. ,. D. R. Jayashree Nair, "An Efficient English to Hindi Machine Translation System Using Hybrid Mechanism," in 2016 International Conference on Advances in Computing, Communications and Informatics (ICACCI), Jaipur, India, 2016.

[12] M. K. S. Shailendra Kumar Singh, "GRT: Gurmukhi to Roman Transliteration System using Character Mapping and Handcrafted Rules," International Journal of Innovative Technology and Exploring Engineering (IJITEE), vol. 8, no. 9, July 2019, 2019.

[13] A. H. M. T. P. S. S. R. Pasindu Tennage, "Transliteration and Byte Pair Encoding to Improve Tamil to Sinhala Neural Machine Translation," in IEEE, Moratuwa, Sri Lanka, 2018.

[14] S. A. Lahiru de Silva, "Singlish to Sinhala Transliteration using Rule-based Approach," in 2021 IEEE 16th International Conference on Industrial and Information Systems (ICIIS), Kandy, Sri Lanka, 2021.

[15] S. H., P. T. V., R. G. Attri, "HiPHET : a hybrid approach for translating code-mixed language (Hinglish) to pure languages (Hindi and English)," Computer Science, vol. T. 21 (3), no. 2020, p. 371–391, 2020.

[16] A. De Silva, "Singlish to Sinhala Converter using Machine Learning," UCSC Digital Library, Colombo, 5-Aug-2021.

[17] M. S. A. S. W. A. @. W. Y. W. M. N. H. N. N. Sitti Munirah Abdul Razak, "Transliteration Engine for Union Catalogue of Malay Manuscripts in Malaysia: E-Jawi Version 3," in Transliteration Engine for Union Catalogue of Malay Manuscripts in Malaysia: E-Jawi Version 3, Kuala Lumpur, Malaysia, 2018